\documentclass[journal,twoside,web]{ieeecolor}

\usepackage{jsen}
\usepackage{url}
\usepackage{algorithm}
\usepackage{algpseudocode}
\usepackage{placeins}
\usepackage{hyperref}

\usepackage{graphicx}
\usepackage{wrapfig}
\usepackage{svg}
\usepackage{hhline}
\usepackage{cite}
\usepackage{amsmath,amssymb,amsfonts}
\usepackage{textcomp}
\usepackage{comment}
\usepackage{adjustbox}
\usepackage{subcaption}
\usepackage{multirow}
\usepackage{booktabs}
\usepackage{color, colortbl}
\usepackage{tikz}
\usetikzlibrary{backgrounds}
\usetikzlibrary{shapes.geometric}
\usetikzlibrary{shapes,arrows,positioning}
\usetikzlibrary{calc}
\usetikzlibrary{arrows.meta, fit}
\usepackage{fontawesome}
\usepackage{makecell}
\usepackage{utfsym}
\usepackage[flushleft]{threeparttable}

\newcommand{\mycirc}[1][black]{\textcolor{#1}{\ensuremath\bullet}}
\usepackage{ifthen}
\usepackage{makecell}

\definecolor{stationary-blue}{RGB}{31, 119, 180}
\definecolor{moving-object-orange}{RGB}{255, 127, 14}
\definecolor{clutter-green}{RGB}{44, 160, 44}

\usepackage{hyperref}
\hypersetup{
    colorlinks=true,
    linkcolor=black,
    filecolor=magenta,      
    urlcolor=cyan,
    citecolor=red,
}

\usepackage{pifont}
\newcommand{\cmark}{\ding{51}}%
\newcommand{\xmark}{\ding{55}}%

\def\BibTeX{{\rm B\kern-.05em{\sc i\kern-.025em b}\kern-.08em    T\kern-.1667em\lower.7ex\hbox{E}\kern-.125emX}}
    
\definecolor{abstractbg}{rgb}{0.89804,0.94510,0.83137}
\definecolor{headerbg}{RGB}{178, 190, 181}

\DeclareMathOperator*{\argmin}{arg\,min}

\def\useblue{0} 
\def\icasspaccepted{1}    

\newcommand{\revcolortext}[1]{%
    \if\useblue1\textcolor{blue}{#1}\else#1\fi%
}

\newcommand{\renderbalancedtable}[1]{%
    \if\useblue1%
        \begin{tikzpicture}[baseline=(tbl.base)]
            \node[inner sep=0, outer sep=0] (tbl) {#1};
            \begin{scope}[on background layer]
                \fill[blue, opacity=0.5] (tbl.north west) rectangle (tbl.south east);
            \end{scope}
        \end{tikzpicture}%
    \else%
        #1
    \fi%
}

\newcommand{\icassptext}[2]{%
    \if\icasspaccepted1#1\else#2\fi%
}

\begin{document}
\title{Lightweight graph transformers for clutter and target classification in automotive radar
}
\author{Adrian Gheorghiu, Tunc Alkanat, Ashish Pandharipande
\thanks{This work was supported in part by the Dutch Ministry
of Economic Affairs and in part by the
Important Projects of Common European Interest (IPCEI)
Microelectronics and Communication Technology (ME/CT)
Project.}
\thanks{Adrian Gheorghiu, Tunc Alkanat, and Ashish Pandharipande are with NXP Semiconductors, The Netherlands (e-mail: adriangheorghiu00@gmail.com; tunc.alkanat@nxp.com; ashish.pandharipande@nxp.com)}
}

\IEEEtitleabstractindextext{%
\fcolorbox{abstractbg}{abstractbg}{%
\begin{minipage}{\textwidth}%
\begin{wrapfigure}[12]{r}{4in}%
\includegraphics[width=4in]{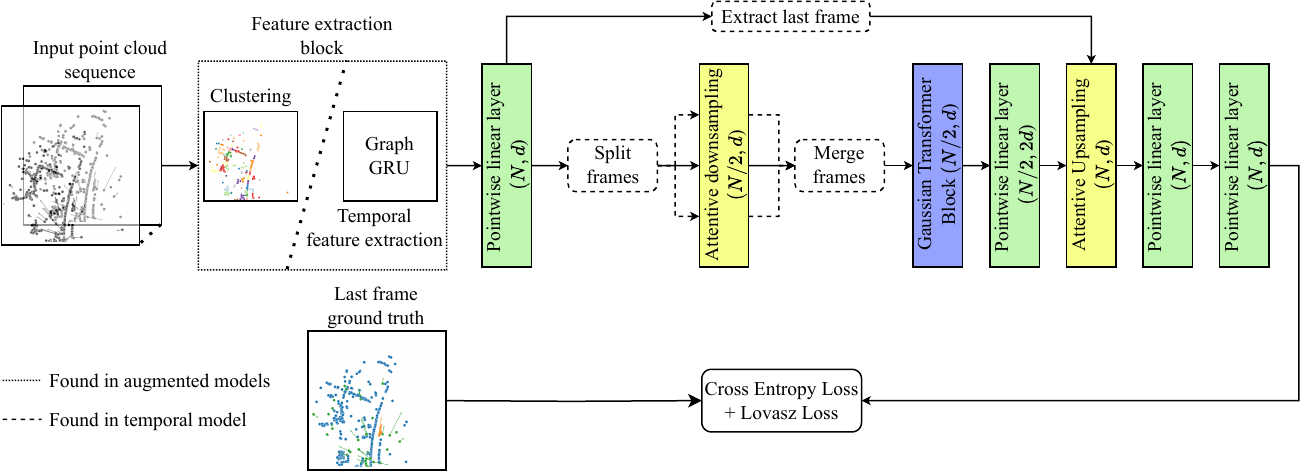}
\end{wrapfigure}%

\begin{abstract}
Radar is a core sensor modality for scene perception to achieve higher levels of autonomy in automotive driving. A common occurrence in automotive radars is clutter - detections of non-existent moving objects, that can adversely impact target detection and classification performance and subsequent driving actions. We propose lightweight tiny graph transformer network (TGTNet) models for classifying clutter from stationary and moving targets in the scene. Performance evaluation on the public RadarScenes show that our proposed TGTNet models achieve similar classification performance in precision and recall metrics in comparison to state of the art models, with one to two orders of magnitude lower model size and significantly faster inference.
\end{abstract}

\end{minipage}}}

\maketitle

\section{Introduction}

Automotive radars play an integral 
role in advanced driver assistance systems (ADAS) and autonomous driving systems (ADS), by providing range, Doppler, and angular information of targets in the driving scene. Their low cost and environmentally robust sensing capabilities make radars a core sensor in various automotive driving functionalities, such as adaptive cruise control, lane assist, and autonomous emergency braking. However, despite their higher robustness in comparison to sensors like cameras and LiDARs, automotive radars are susceptible to signal artifacts, such as clutter.

\begin{figure}[!ht]
\centering

\includegraphics[width=0.7\linewidth]{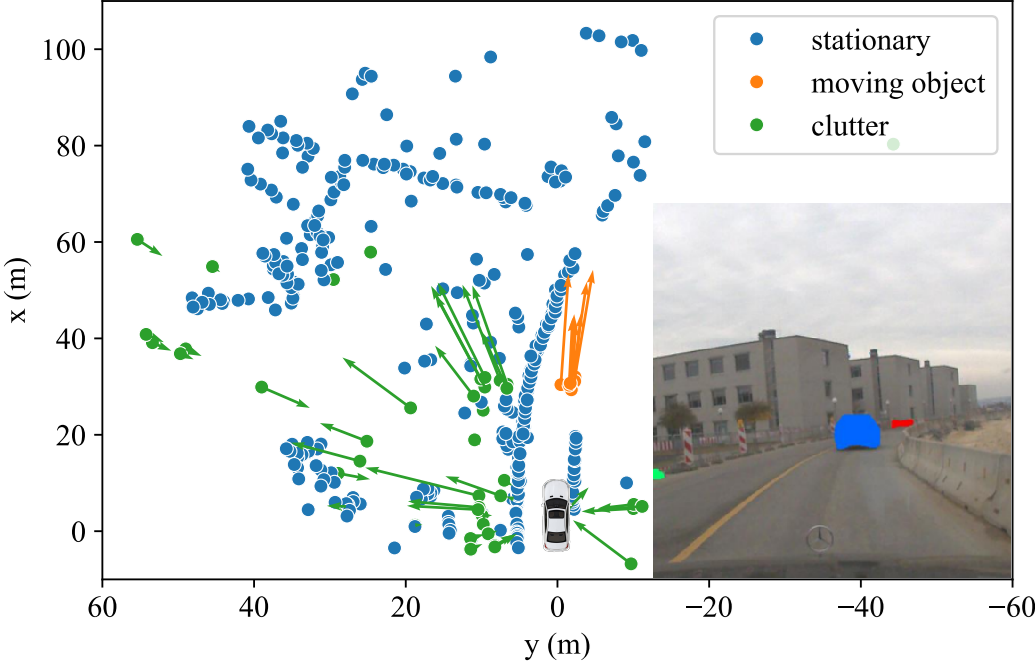}
\caption{Illustration depicting radar point cloud with clutter, stationary, and moving targets.}
\label{fig:clutter}
\end{figure}

Radar clutter refers to erroneous detections in the radar system due to a non-existent object. As depicted in Fig.~\ref{fig:clutter} for an illustrative scene, clutter can occur due to indirect signal propagation paths between the target and the radar, or erroneous ambiguity resolution in the radar processing chain. Specifically, ghost detections are a subclass of clutter detections that are caused by multipath signal reflections. These ghost targets often occur when radar signals bounce off large reflective surfaces such as buildings, metallic guardrails, other vehicles, or in environments like tunnels, and then arrive at the radar receiver via indirect paths  \cite{RuleBasedClutterDetection, GhostDataset}.

Clutter detections impact the overall radar detection performance and can mislead decision-making processes, e.g., leading to faulty braking or steering maneuvers. Hence, robust detection and mitigation of clutter targets are essential for highly reliable radar sensing and perception.

In this work, we propose \textbf{t}iny \textbf{g}raph \textbf{t}ransformer \textbf{net}work (TGTNet) models to classify clutter, stationary, and moving targets using radar point cloud data. The first class of TGTNet models use a single point cloud frame, while the second class uses temporal augmentation.
Our main contributions are:

\noindent \textbf{i.} We propose TGTNet, lightweight point transformer models for efficient radar point cloud clutter segmentation.
    We propose the Superpoint Block, a point cloud neural network module that leverages energy model clustering to extract latent features.
    Model size is reduced further by employing a low-rank vector attention mechanism.

\noindent \textbf{ii.} Two temporal augmentations are incorporated in TGTNet to exploit time-series point cloud information.

\noindent \textbf{iii.} All proposed models are evaluated against state of the art models using the RadarScenes  public dataset that contains clutter and target annotations \cite{RadarScenes, GhostLabeledRadarScenes}. Qualitative and quantitative results are presented to demonstrate that the proposed models show comparable classification performance while having one to two orders of magnitude lower model sizes. 

\section{Related literature}

One class of works for ghost target detection is based on signal processing approaches.
A simple model for ghost target detection based on the deviation of the angle of the motion vector from a box model was presented in \cite{RSB}. The existence of multipath was modeled as a binary composite hypothesis test and a generalized likelihood ratio test detector was proposed in \cite{ZLL} to determine whether indirect paths exist in a range-Doppler cell. Both of these theoretical approaches were evaluated with limited simulations. In \cite{RuleBasedClutterDetection}, the authors observe that ghost detections are inherently a geometrical phenomenon, and propose a simple rule-based method for detecting clutter by employing searches within the detections at the current timestep and a buffer of the previous timesteps. 

A multistage rule-based multipath ghost object labeling technique leveraging LiDAR is proposed in \cite{DB}.
Region proposals of multipath ghost reflections are created by reflecting real objects off reflective surfaces, and then candidate ghost reflections are chosen based on a Mahalanobis distance threshold. Performance is reported on the Radar Ghost dataset \cite{GhostDataset}.

As data-driven point cloud processing methods gained traction, deep learning methods for clutter handling have emerged. The Ghost Radar Dataset in \cite{GhostDataset} is currently the only publicly available dataset that provides manually labeled ghost objects. The annotations are, however, limited to a single type of multipath reflection that is caused by a single reflective surface. Along with the dataset, two instance segmentation methods were presented that use PointNet\texttt{++} \cite{PointNet++} as the semantic segmentation backbone. The first method utilizes a similarity group proposal network \cite{SGPN} in an end-to-end training scenario, while the second method applies the well-known DBSCAN algorithm \cite{DBSCAN} to the semantic segmentation outputs.


In \cite{RadarMLPhDThesis}, a modified PointNet\texttt{++} \cite{PointNet++} variant with two abstraction levels was proposed for moving object segmentation. Training and testing were done on the RadarScenes dataset \cite{RadarScenes} by aggregating points in a 500~ms time window. The architecture was designed to process point clouds with a fixed size of 3072 points. If the point cloud has fewer points, it is oversampled; if it has more points, it is undersampled by selecting the points with the highest Doppler values.

An automatically labeled public dataset with clutter detections was proposed in \cite{GhostLabeledRadarScenes}, along with a modified deep learning architecture for ghost object detection. The automatic labeling method was applied to the RadarScenes dataset, which provided annotations for moving objects. First, the background detections found in close proximity to the ``real'' moving object were also labeled as ``real'' to account for radar imprecision. All the remaining detections that have the ego-motion-compensated radial speed above a certain threshold were marked as ``clutter''. The authors then proposed a modified version of the PointNet\texttt{++} architecture from \cite{RadarMLPhDThesis} and tested it on both the new dataset and the private dataset from their previous work \cite{RuleBasedClutterDetection} to validate the label generation method. The model input was reduced to a smaller fixed size of 1280 points. Point clouds were aggregated in a queue of fixed size, with the oldest points discarded when new points come in. The point clouds were then brought to the same coordinate space using ego motion information, and training was done by oversampling points such that every sample in the batch had the same number of points. This method and the one in \cite{RadarMLPhDThesis} are both specifically tailored to radar point clouds, making them suitable baselines for our models.

The Point Transformer was proposed in \cite{PointTransformer}. The Gaussian Radar Transformer (GRT), a new point transformer segmentation model, was presented in \cite{GaussianRadarTransformer}, employing a decoupled point attention mechanism that facilitates the flow of gradients. To enhance the model's ability to learn spatial relations, attentive downsampling and upsampling layers were proposed, utilizing a simplified translation-sensitive attention mechanism that eliminates the need for key, query, and value separation. The model was trained and tested without any temporal aggregation by merging frames from all sensors into a single point cloud. Using this architecture, state-of-the-art performance was achieved on the RadarScenes moving object segmentation task.


Generally, deep learning approaches to clutter detection in literature use variations of PointNet\texttt{++} with various methods for point cloud aggregation, without consideration to model complexity which is an important aspect for automotive edge implementations. In this work, we\icassptext{ extend our previous work~\cite{gheorghiu2026tgtnet} and}{} develop models that take advantage of recent advances in graph neural networks and transformer architectures that have thus far been  overlooked. 
\icassptext{We efficiently extend our models proposed in \cite{gheorghiu2026tgtnet} to time domain capabilities, and present additional qualitative visualizations as well as new quantitative results on a second dataset, the Radar Ghost Dataset \cite{GhostDataset}, showcasing the wider applicability of our models.}{}


\label{sect:intro}


\section{Proposed lightweight transformer models}
\label{sect:method}

Point Transformers \cite{PointTransformer} are characterized by their application of an attention mechanism on localized structures which are defined by neighborhoods within the point cloud. Given an unordered point cloud $\mathcal{P}$ consisting of Cartesian positions $p \in \mathbb{R}^{N \times c}$ and endowed with features $x \in \mathbb{R} ^ {N \times d}$, a Point Transformer uses a learnable function $\theta$ to process the relative position between a point and its neighbors and extract a positional encoding 
\begin{equation}
    \delta = \theta(p_i - p_j)
\end{equation}
that is used in the attention mechanism.

The scaled dot-product attention of the original Transformer is changed in favor of vector attention. As a result, the Point Transformer attention layer is defined as
\begin{equation}
    h_i = \sum_{j \in \mathcal{N}(i)}\rho \left(\gamma \left(\varphi(x_i) - \psi(x_j) + \delta \right) \right) \odot \left(\alpha(x_j) + \delta \right) \mathrm{,}
\end{equation}
where $\varphi$, $\psi$, and $\alpha$ are linear projections of the query, key, and value, respectively~\cite{PointTransformer}. $\gamma$ is a learnable mapping function (e.g., a multi-layer perceptron, MLP), $\rho$ is a normalization function (e.g., softmax), and $\mathcal{N}$ is a neighborhood function.

For our architecture, we take inspiration from \cite{GaussianRadarTransformer} and use the Gaussian activation function
\begin{equation} \label{eq:gaussian_activation}
    G(x)=\exp\left(\frac{-x^2}{2}\right)
\end{equation}
as the normalization function, for its property of facilitating gradient flow. Additionally, this activation function facilitates the identification of similarities between points, as the attention weights between dissimilar features will be weighed down, while weights between similar features will be larger. This choice improves the expressivity of our small models.

To facilitate learning long-range dependencies for isolated points, the neighborhood function is chosen as the k-nearest neighbors (kNN). The attention mechanism is also simplified to exclude the positional encoding for the values, and the additional mapping $\gamma$ is discarded. The output is then passed through an output projection with bias $\omega$, resulting in the Gaussian Attention Layer
\begin{equation}\label{eq:gtb_attention}
    h_i = \omega \left( \sum_{j \in \operatorname{kNN}(i)}G \left(\varphi(x_i) - \psi(x_j) + \delta \right) \odot \alpha(x_j) \right) \mathrm{,}
\end{equation}
which is followed by a GELU activation \cite{GELU} and placed between two pointwise linear layers, also followed by GELU activations, forming the Gaussian Transformer Block (GTB). The output projection $\omega$ concentrates the bias of the attention layer into one single projection and serves as a post-processing step for the attention mechanism.

Most clutter detections are found behind real objects and are caused by reflections of the real detections. These properties make the vector attention with element-wise feature subtraction found in the GTB suitable for learning feature dependencies and similarities that are indicative of clutter.

The MLP $\theta$ that extracts the positional encoding $\delta$ from the relative position uses two linear layers with a batch normalization layer \cite{BatchNorm} applied across all encodings in the batch. The first linear layer keeps the dimension of the positions $c$, and the output is passed through the batch normalization layer and a ReLU activation. The second linear layer projects the positional encoding to the dimension $d$ of the attention layer.

\subsection{Radar point cloud clustering}





\begin{figure}[!ht]
\centering

\includegraphics[width=0.67\linewidth]{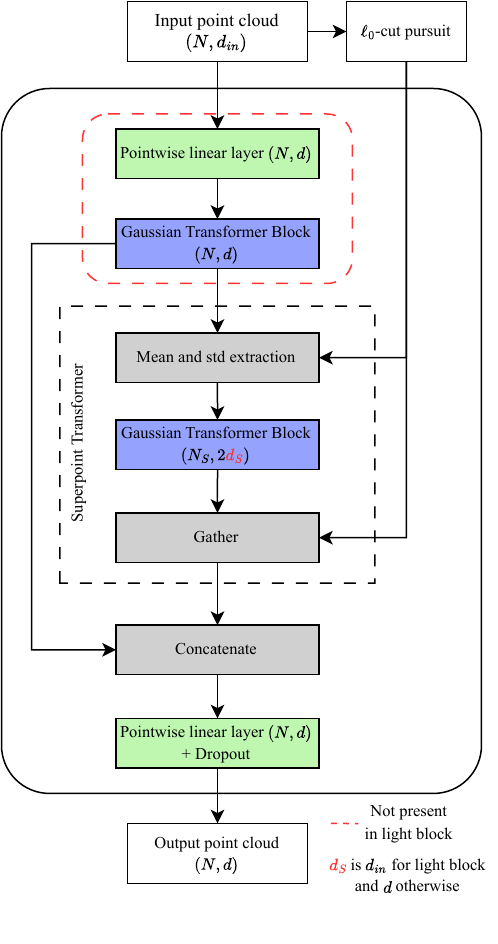}

\caption{Diagram of the regular and light Superpoint Blocks. The mean and standard deviation are extracted for both the positions and features, with the Superpoint Transformer operating on a point cloud with four positional dimensions. The pointwise linear layers are followed by ReLU activations.}

\label{fig:superpoint_block}

\end{figure}

\begin{figure*}[ht]
\centering

\includegraphics[width=0.8\linewidth]{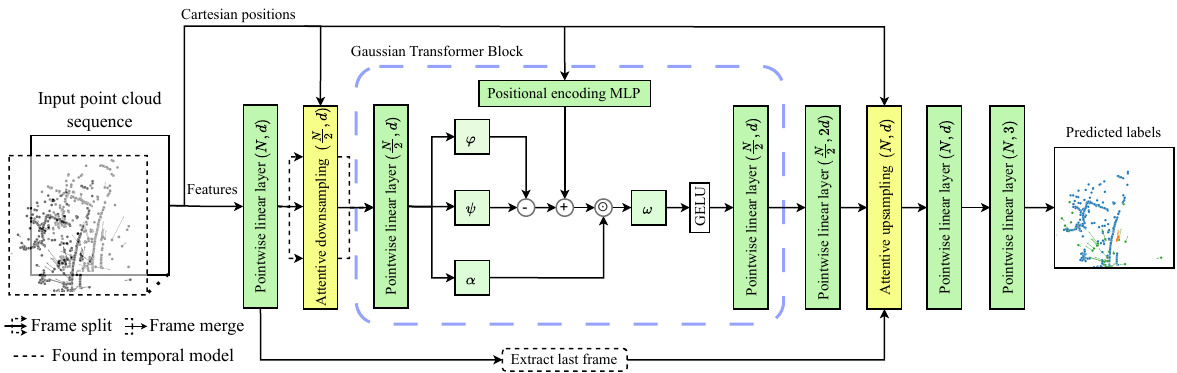}

\caption{Diagram of TGTNet. All pointwise linear layers except the last one are followed by GELU activations.}
\label{fig:base_model}

\end{figure*}

To facilitate learning local features, we additionally introduce a new module, the Superpoint Block, which extracts features from a clustered point cloud input. The base building block of the Superpoint Blocks is the Superpoint Transformer. The Superpoint transformer takes as input a clustered point cloud and extracts the mean and standard deviation of the positions and features, resulting in $N_S$ superpoint positions $[\bar{p}_s, p_s^{(\mathrm{std})}]$ and $N_S$ superpoint features $[\bar{x}_s, x_s^{(\mathrm{std})}]$, where $s$ represents a unique cluster index. The superpoint positions and features extracted from the clusters are then treated as a point cloud and passed through a GTB. Finally, the output superpoints are applied a gather operation, which assigns to each original point the relevant output superpoint that represents the same cluster to which the point belongs. Used in conjunction with a skip connection, the Superpoint Transformer extracts useful features for each point based on the cluster mean and spread as well as the relations to nearby clusters of similar spread. Based on the Superpoint Transformer, two Superpoint Block versions are then proposed, as seen in Fig. \ref{fig:superpoint_block}.

The first block version applies a pointwise linear layer that first projects the features to the block dimension, followed by a GTB. The output of the GTB is then passed through a Superpoint Transformer. The features of the two outputs are concatenated with the output positions of the Superpoint Transformer, then passed through a pointwise linear layer that reduces the feature dimension to the block dimensions, and finally passed through a dropout layer.

The second block is a lighter version of the first block that forgoes the initial linear layer and GTB. Other than that, the features and positions are passed through the Superpoint Transformer exactly like the first module.

To cluster the point clouds, we branch out from the pervasive DBSCAN algorithm \cite{DBSCAN} and its variants, which are often used in radar point cloud clustering. In that sense, we partition the point cloud using a global energy model similar to the one used in \cite{SuperpointGraphs}. Therefore, the energy model is defined on top of the kNN of the point cloud and considers the Cartesian positions as a signal $f \in \mathbb{R}^{N \times c}$. The objective is to find a piecewise constant signal $g$ that approximates $f$ by minimizing the energy model
\begin{equation}
    g^{\star}= \argmin_{g \in \mathbb{R}^{N \times c}} \sum_{i=1}^N||g_i - f_i|| ^2 + \lambda \sum_{i=1}^N\sum_{j \in \operatorname{kNN}(i)} [g_i \neq g_j] \mathrm{,}
\end{equation}
where the $[\cdot]$ operator is known as the Iverson bracket. This energy model consists of two parts: one that ensures fidelity to the original signal through the squared Euclidean norm, and a regularizer with regularization strength $\lambda$ that determines the number of distinct values in the signal. The partition is then defined as the connected components of the kNN graph for which the approximated signal $g^{\star}$ is constant.
Despite the direct minimization of the energy model not being computationally tractable, it can be efficiently approximated by the iterative $\ell_0$-cut pursuit algorithm in \cite{cutpursuit}.

Our approach diverges from DBSCAN due to its sensitivity to hyperparameter selection and its limitations when handling clusters with varying densities. Additionally, DBSCAN often classifies isolated detections as noise, resulting in the loss of potentially valuable information. In contrast, the proposed energy-based clustering method is specifically designed to accommodate the sparsity and irregularity of radar point clouds. This, in turn, ensures that our Superpoint Block learns important local relations between elements with simple shapes. To augment TGTNet with the Superpoint Blocks, the blocks are prepended to TGTNet. The models augmented with the normal and light versions of the Superpoint Block are then called S-TGTNet and SL-TGTNet, respectively.

\subsection{Single snapshot models}

\begin{table*}
\centering
\caption{Comparison of our TGTNet model for different sizes with a baseline.}
\resizebox{\linewidth}{!}{
\begin{tabular}{cccccccccccccccc}
\hline
\multirow{2}{*}{\textbf{Setup}} & \multirow{2}{*}{\textbf{LoRa}} & \multirow{2}{*}{\textbf{Model size}} & \multirow{2}{*}{\textbf{\textbf{\begin{tabular}[c]{@{}c@{}}Inf. time\\ (ms)\end{tabular}}}} & \multicolumn{3}{c}{\textbf{Moving object}}                      & \multicolumn{3}{c}{\textbf{Clutter}}                            & \multicolumn{3}{c}{\textbf{Stationary}}                         & \multicolumn{3}{c}{\textbf{Mean}}                               \\
                                &                                &                                      &                                               & Precision & \multicolumn{1}{c}{Recall} & \multicolumn{1}{c}{F1} & Precision & \multicolumn{1}{c}{Recall} & \multicolumn{1}{c}{F1} & Precision & \multicolumn{1}{c}{Recall} & \multicolumn{1}{c}{F1} & Precision & \multicolumn{1}{c}{Recall} & \multicolumn{1}{c}{F1} \\ \hline
GRT \cite{GaussianRadarTransformer}      & -                              & 20.07 MB                             & 25.47                                         & 90.76     & 91.47                      & 91.12                  & 95.42     & 95.32                      & 95.37                  & 99.88     & 99.85                      & 99.86                  & 95.35     & 95.55                      & 95.45                  \\ \hline
TGTNet (dim 32)                 & \xmark                              & 80.25 KB                             & 6.83                                          & 89.49     & 88.81                      & 89.15                  & 94.50     & 94.27                      & 94.38                  & 99.82     & 99.86                      & 99.84                  & 94.60     & 94.32                      & 94.46                  \\ \hline
TGTNet (dim 32)                 & \cmark                              & 40.29 KB                             & 7.10                                          & 88.33     & 89.29                      & 88.81                  & 94.66     & 93.85                      & 94.25                  & 99.83     & 99.83                      & 99.83                  & 94.28     & 94.33                      & 94.30                  \\ \hline
TGTNet (dim 16)                 & \xmark                              & 21.19 KB                             & 6.93                                          & 85.39     & 87.49                      & 86.43                  & 93.70     & 92.05                      & 92.87                  & 99.79     & 99.78                      & 99.79                  & 92.96     & 93.11                      & 93.03                  \\ \hline
TGTNet (dim 16)                 & \cmark                              & 11.21 KB                             & 7.39                                          & 85.33     & 87.15                      & 86.23                  & 93.21     & 91.95                      & 92.58                  & 99.79     & 99.77                      & 99.78                  & 92.78     & 92.96                      & 92.86                  \\ \hline
\end{tabular}
}
\label{table:results_base}
\end{table*}

Our architecture, the Tiny Graph Transformer Network (TGTNet), takes the form of a simplified U-Net \cite{UNet} with only one level of abstraction. The architecture is designed with translation invariance in mind and can operate on point clouds of varying sizes. Therefore, the input point cloud is passed through a pointwise linear layer, then through a downsampling layer. The downsampled point cloud is passed through a Gaussian Transformer Block (GTB) and another pointwise linear layer before being upsampled again through the upsampling block with a skip connection from the first linear layer. The upsampled point cloud is finally passed through a pointwise classification head MLP with two layers.


To downsample the point cloud, we employ the farthest-point sampling (FPS) algorithm, first proposed for use on point clouds in PointNet++ \cite{PointNet++}, and set the downsampling ratio to halve the number of points after downsampling. To facilitate learning useful features for the downsampled point clouds, we again take inspiration from \cite{GaussianRadarTransformer} and their attentive downsampling layer. The attentive downsampling layer presented in \cite{GaussianRadarTransformer} uses a simplified attention mechanism that operates on the downsampled point cloud by aggregating the features of the 9-nearest neighbors in the original point cloud. Instead of using three separate matrices for extracting query, key, and value representations, the authors draw inspiration from \cite{SimplifiedPointAttention} and process the features using a shared pointwise MLP. They then calculate the weights by applying a second MLP to the processed weights, concatenated with the Cartesian positions.

Due to the lack of positional invariance of the aforementioned attentive downsampling layer combined with the high parameter count from the use of two input MLPs, we propose our own attentive downsampling layer. To this end, our layer reverts back to the full attention in  (\ref{eq:gtb_attention}) and applies it on the downsampled point cloud by aggregating points from the original point cloud. In order to introduce more knowledge about the local regions of the downsampled points, the attention layer output for each point is concatenated with the standard deviation of the Cartesian coordinates in the aggregated region before being multiplied by the output projection $\omega$.

The upsampling layer uses the same idea as the downsampling layer. This time, the query features originate from the skip connection of the first pointwise linear layer, while the key and value features originate from the downsampled point cloud. The only other difference is that the resulting queries, keys, and values must share the same dimension; however, the points they originate from have different feature dimensions, so a common dimension needs to be chosen for the features to be projected to. As a result, we decided to project everything to the dimensions of the skip connections.

To further decrease the memory footprint of our model, we also test modified versions of the attention layer and Gaussian Transformer Block architectures that learn low-rank (LoRa) adaptations for a subset of the linear layers. Inspired by LoRa adaptations \cite{LoRa} and the recent advancements in LLMs brought by DeepSeek-v3 \cite{DeepSeek}, we propose the use of learning compressed representations in our attention layers and label the models that use this attention as ``LoRa'' models.

To this end, the LoRa variants of the base models have all attentive layers modified, while augmentation blocks used for models have their own separate LoRa qualifier. For the Gaussian Transformer Block and our proposed attentive downsampling and upsampling layers, the linear projection matrices in (\ref{eq:gtb_attention}) that extract the query, key, and value are transformed into low-rank projections. As a result, a matrix $W \in \mathbb{R}^{d_1 \times d_2}$ is transformed into two subsequent projection matrices $A\in \mathbb{R}^{k \times d_2}$ and $B \in \mathbb{R}^{d_1 \times k}$, with $k$ being the compressed dimension. Furthermore, the keys and values are extracted using the same shared compression matrix, but different decompression matrices. In the case of the Gaussian Transformer Block, the linear layers that encapsulate the attention layer are also transformed into low-rank projections.


In all cases, we empirically chose the compressed dimension to be the input dimension divided by 8. For layers that process input features of different dimension sizes, we chose the larger input dimension of the two. 

\subsection{Time-series snapshot models}

\begin{table*}[!ht]
\centering
\caption{Comparison of our Superpoint Block models. Best values are shown in \textcolor{red}{red}, second best in \textcolor{blue}{blue}. The reduced feature set is $x=[x_{cc}, y_{cc}, \hat{v}_{r,sc}, \sigma]$, while the extended feature set is $x=[x_{cc}, y_{cc}, \hat{v}_{r,sc}, \sigma, r_{sc}, \phi_{sc}, t, s_{id}]$.}
\resizebox{\linewidth}{!}{
\begin{threeparttable}
\begin{tabular}{cccccccccccccccc}
\hline
\multirow{2}{*}{\textbf{Setup}} & \multirow{2}{*}{\textbf{LoRa\textsuperscript{\textdagger}}} & \multirow{2}{*}{\textbf{Model size}} & \multirow{2}{*}{\textbf{\begin{tabular}[c]{@{}c@{}}Inf. time\\ (ms)\end{tabular}}} & \multicolumn{3}{c}{\textbf{Moving object}} & \multicolumn{3}{c}{\textbf{Clutter}} & \multicolumn{3}{c}{\textbf{Stationary}} & \multicolumn{3}{c}{\textbf{Mean}} \\
                                &                                                                                           &                                      & \multicolumn{1}{l}{}                                              & Precision       & Recall      & F1         & Precision     & Recall    & F1       & Precision      & Recall     & F1        & Precision    & Recall   & F1      \\ \hline
S-TGTNet* (dim 16)              & \xmark                                                                                         & 47.11 KB                             & 11.99                                                             & 90.32           & 90.05       & 90.18      & 94.82         & 94.76     & 94.79    & 99.85          & 99.87      & \textcolor{blue}{99.86}     & 95.00        & 94.89    & 94.94   \\ \hline
SL-TGTNet* (dim 16)             & \xmark                                                                                         & \textcolor{blue}{14.88 KB}                             & 10.56                                                             & 88.85           & 85.63       & 87.21      & 92.95         & 93.20     & 93.07    & 99.75          & 99.87      & 99.81     & 93.85        & 92.90    & 93.36   \\ \hline
S-TGTNet* (dim 16)              & \cmark                                                                                         & 27.78 KB                             & 12.80                                                             & 90.11           & 89.04       & 89.57      & 94.27         & 94.91     & 94.59    & 99.85          & 99.86      & 99.85     & 94.74        & 94.60    & 94.67   \\ \hline
SL-TGTNet* (dim 16)             & \cmark                                                                                         & \textcolor{red}{13.92 KB}                             & 11.10                                                             & 87.56           & 88.82       & 88.18      & 94.74         & 90.62     & 92.63    & 99.70          & 99.86      & 99.78     & 94.00        & 93.10    & 93.53   \\ \hline
S-TGTNet (dim 16)               & \cmark                                                                                         & 28.03 KB                             & 12.01                                                             & \textcolor{blue}{90.92}           & 89.39       & 90.15      & 94.80         & 92.67     & 93.72    & 99.71          & \textcolor{red}{99.89}      & 99.80     & 95.14        & 93.98    & 94.56   \\ \hline
SL-TGTNet (dim 16)              & \cmark                                                                                         & 16.52 KB                             & \textcolor{red}{10.12}                                                             & 86.86           & \textcolor{blue}{90.13}       & 88.47      & 95.37         & 92.57     & 93.95    & 99.82          & 99.82      & 99.82     & 94.01        & 94.17    & 94.08   \\ \hline
S-TGTNet (dim 32)               & \cmark                                                                                         & 103.03 KB                            & 12.49                                                             & \textcolor{red}{90.97}           & \textcolor{red}{91.49}       & \textcolor{red}{91.23}      & \textcolor{red}{95.82}         & \textcolor{blue}{95.06}     & \textcolor{red}{95.44}    & \textcolor{blue}{99.86}          & \textcolor{blue}{99.88}      & \textcolor{red}{99.87}     & \textcolor{red}{95.55}        & \textcolor{red}{95.48}    & \textcolor{red}{95.51}   \\ \hline
S-TGTNet\textsuperscript{\#} (dim 32) & \cmark                                                                                         & 103.03 KB                            & 12.49                                                             & 90.75           & 91.09       & \textcolor{blue}{90.92}       & \textcolor{blue}{95.44}         & \textcolor{red}{95.25}     & \textcolor{blue}{95.35}    & \textcolor{red}{99.87} & 99.86      & \textcolor{blue}{99.86}     & \textcolor{blue}{95.35}        & \textcolor{blue}{95.40}    & \textcolor{blue}{95.38}   \\ \hline
SL-TGTNet (dim 32)              & \cmark                                                                                         & 50.16 KB                             & \textcolor{blue}{10.44}                                                             & 89.05           & 89.95       & 89.50      & 95.15         & 92.48     & 93.80    & 99.76          & 99.86      & 99.81     & 94.65        & 94.10    & 94.37   \\ \hline
\end{tabular}
\end{threeparttable}
}
\begin{tablenotes}
    \item * Reduced feature set.
    \quad \revcolortext{\textsuperscript{\#} DBSCAN clustering}
    \quad \textsuperscript{\textdagger} Refers to Superpoint Block. All base models are LoRa.\vspace{-0.5cm}
\end{tablenotes}
\label{table:results_superpoints}
\end{table*}

To improve the performance of our small, optimized models and make them comparable to larger models, we test two methods for incorporating multiple radar frames as input.

First, we test temporal feature extraction using the PointGRU method presented in \cite{pointrnn}. For this, we adapt the PointGRU equations to utilize the point transformer attention as the base building block for computing the update and reset gate point clouds, as well as the candidate point cloud. Due to experiencing divergence during training when using Gaussian normalization for the attention coefficients, we use standard softmax normalization for the PointGRU unit instead. Additionally, since temporal relations between points happen across small spatial distances, the kNN neighborhood is switched for the radius neighborhood:
\begin{equation}\label{eq:point-attention}
    h_i = \sum_{j \in \operatorname{radius}(i)}\operatorname{softmax} \left(\varphi(x_i) - \psi(x_j) + \delta \right) \odot \alpha(x_j) \mathrm{.}
\end{equation}
The $\operatorname{point-attention-rnn}$ is then defined by applying (\ref{eq:point-attention}) to every point in the current point cloud $\mathcal{P}_t=(P_t, X_t)$ by aggregating the points in the current point cloud and the point cloud of the previous state $\mathcal{S}_{t-1}=(P_{t-1}, S_{t-1})$. As a result, the PointGRU equation uses three different $\operatorname{point-attention-rnn}$ layers, each with its own set of parameters:
\begin{equation}
    Z_t=\sigma(\operatorname{point-attention-rnn_{u}}((P_t, X_t), (P_{t-1}, S_{t-1}))) \mathrm{,}
\end{equation}
\begin{equation}
    R_t=\sigma(\operatorname{point-attention-rnn_{r}}((P_t, X_t), (P_{t-1}, S_{t-1}))) \mathrm{,}
\end{equation}
\begin{equation}\label{eq:point_gru_prev_registered}
    \hat{S}_{t-1}=\operatorname{point-attention-rnn_{s}}((P_t, X_t), (P_{t-1}, S_{t-1})) \mathrm{,}
\end{equation}
\begin{equation}
    \tilde{S}_{t}=\tanh(W_{\tilde{s}}[X_t, R_t \odot \tilde{S}_{t-1} ] + b_{\tilde{s}}) \mathrm{,}
\end{equation}
\begin{equation}
    S_t = Z_t \odot \tilde{S}_{t-1} + (1 - Z_t) \odot \tilde{S}_t \mathrm{.}
\end{equation}
Here, $\sigma$ is the sigmoid function and $\tanh$ is the hyperbolic tangent function. This layer is used to extract temporally rich features, which are then passed onto our model to segment the last frame's point cloud, an architecture which we call GRU-TGTNet. It is worth noting that  (\ref{eq:point_gru_prev_registered}) also deviates from the original paper \cite{pointrnn}, and is changed to additionally take the features of the current point cloud $X_t$ as input. This modification is made to condition the GRU twice on the current point cloud, thus helping reinforce the importance of the most recent point cloud.

Second, we test a simpler method for incorporating the time domain, TGTNet-T, which involves including the index of the frame number in the sequence as a feature. This method processes all frames independently in the downsampling part, then aggregates them to the same coordinate space before they are passed through the Gaussian Transformer Block. Finally, for upsampling, only the point cloud of the last frame is passed as a skip connection.

\section{Performance evaluation}
\label{sect:exp}

\subsection{Dataset and evaluation metrics}

Our models were trained and evaluated on the RadarScenes dataset \cite{RadarScenes}; the method in \cite{GhostLabeledRadarScenes} was used to for ground truth class labels. We utilize the original dataset split from \cite{RadarScenes} in conjunction with the additional test and validation split from \cite{GaussianRadarTransformer}. Specifically, sequences 6, 42, 58, 85, 99, and 122 are assigned to the validation set, while the remaining sequences from the original validation set are used for the test set. The RadarScenes dataset comprises radar scans from four frequency modulated continuous waveform (FMCW) radar sensors positioned on either side of the car's front bumper, covering different angles. To guarantee that a single radar snapshot captures the entire field of view of the car, the scans were merged from all four sensors in a single frame. 
The point features we trained our models on are the Cartesian positions in car coordinates, $x_{cc}, y_{cc}$, the ego-motion compensated radial velocity, $\hat{v}_{r,sc}$, and the radar cross-section (RCS), $\sigma$. Therefore, a point comprises of position $p=[x_{cc}, y_{cc}]$ and features $x=[x_{cc}, y_{cc}, \hat{v}_{r,sc}, \sigma]$. Some models are also tested using the extended feature set used in \cite{GhostLabeledRadarScenes}, which additionally contains the distance $r_{sc}$ and azimuth $\phi_{sc}$ in sensor coordinates, the timestamp $t$ in seconds relative to the latest frame, and the sensor id $s_{id}$, resulting in $x=[x_{cc}, y_{cc}, \hat{v}_{r,sc}, \sigma, r_{sc}, \phi_{sc}, t, s_{id}]$. The choice of including the Cartesian positions into the features makes the models lose their translation invariance property, but improves performance.

We additionally train and report the performance of our models on the Radar Ghost Dataset, for which we use features $x=[x_{cc}, y_{cc}, \hat{v}_{r,sc}, A, r_{sc}, \phi_{sc}]$, where $A$ is the radar amplitude in dB.

\subsubsection{Model and training details}

For training, we chose to train all models for 50 epochs on RadarScenes and 30 epochs on the Radar Ghost Dataset, and then select the model with the highest F1 score for clutter. 
For our tests, we employ a combination of the Lovász-Softmax loss \cite{LovaszLoss} and a weighted cross-entropy loss. The Lovász-Softmax loss is calculated by applying the Lovász continuous extension to the Jaccard index, and using the resulting extension on the probability distribution output of the model. 
Therefore, for predicted labels $\hat{y}$ and ground truth labels $y$, our loss is defined as
\begin{equation}
    \mathcal{L}(\hat{y}, y)= \mathcal{L}_{Lov\acute{a}sz}(\hat{y}, y) + \mathcal{L}_{CE}(\hat{y}, y, w_c) \mathrm{,}
\end{equation}
where $w_c$ is the vector of the class weights. For RadarScenes, this vector is  set empirically to 0.5, 8, and 8 for the stationary, moving object, and clutter classes, respectively. For the Radar Ghost Dataset, we employ the method described in \cite{BalancedLoss} and calculate the weights based on the effective number of samples. Due to the Lovász-Softmax loss being a tractable surrogate for the mean intersection-over-union (mIoU) metric, we apply it only to clutter and moving object detections to prevent large gradients caused by background class errors in the early stages of training. Moreover, the loss is applied to each individual point cloud from the minibatch and then averaged, thereby ensuring the sample mIoU is maximized. For optimization, we use the Adam optimizer \cite{adam} with a learning rate of 0.0005 and a batch size of 32. The learning rate is reduced after each epoch according to a cosine annealing scheduler \cite{CosineAnnealing} with no restarts.

For our model, we tested two dimensions of size 16 and 32. The overall dimension is kept by the downsampling layer and the Gaussian Transformer Block, while the pointwise layer that follows the transformer block doubles it, and the upsampling layer reduces it back down to its original size. 
For the temporal models, we standardize the positions to the $[-1, 1]$ range and normalize the other features by subtracting their training set mean and dividing by the standard deviation. The Cartesian positions are normalized together so as not to skew the neighborhood computations. The kNN neighborhoods that the layers operate on are set to $k=9$ for the upsampling and downsampling layers and $k=16$ for the Gaussian transformer block layers, which are the same parameters used in \cite{GaussianRadarTransformer}.

For the superpoint blocks, we chose the block dimension to be the same as the dimension of the base model it augments, thus 16 or 32. The clustering hyperparameters are selected to facilitate the extraction of geometric primitives while ensuring the number of superpoints is kept to a minimum. Therefore, we settled on the kNN number of neighbors to be $k=10$ and the regularization strength $\lambda=1$. The Superpoint block output dropout probability is set to 0.1.

For the temporal augmentations, we chose to use six frames for all methods. The GRU uses a dimension of 16 for the embeddings along with a radius of 5 for the neighborhood function, limited to at most 32 neighbors.


To evaluate model performance, we report precision, recall and F1 metrics for the entire test set for all of the classes. \revcolortext{The reported mean F1 scores are macro-averaged, treating all classes with equal importance regardless of dataset imbalance.} 
We also compare the inference time of the models, as well as their size (a small memory footprint is desired for real-time inference in ADAS \cite{ADASDeepLearning}), which is reported for 32-bit floating point parameters.
When measuring inference time, all models are tested under the same conditions and with the same input point cloud, taking into account the frame aggregation method employed by each model. Using a single RTX A6000 GPU, a total of 100 runs are averaged to compute the inference time.

\subsection{Discussion of results}
\label{sect:disc}

\begin{table*}[!ht]
\centering
\caption{Comparison of our temporal methods with baselines. Best values are shown in \textbf{bold}. The reduced feature set is $x=[x_{cc}, y_{cc}, \hat{v}_{r,sc}, \sigma]$, while the extended feature set is $x=[x_{cc}, y_{cc}, \hat{v}_{r,sc}, \sigma, r_{sc}, \phi_{sc}, t, s_{id}]$.}
\resizebox{\linewidth}{!}{
\begin{threeparttable}
\begin{tabular}{cccccccccccccccc}
\hline
\multirow{2}{*}{\textbf{Setup}}             & \multirow{2}{*}{\textbf{LoRa\textsuperscript{\textdagger}}} & \multirow{2}{*}{\textbf{Model size}} & \multirow{2}{*}{\textbf{\begin{tabular}[c]{@{}c@{}}Inf. time\\ (ms)\end{tabular}}} & \multicolumn{3}{c}{\textbf{Moving object}}       & \multicolumn{3}{c}{\textbf{Clutter}}             & \multicolumn{3}{c}{\textbf{Stationary}}          & \multicolumn{3}{c}{\textbf{Mean}}                \\
                                            &                       &                                      & \multicolumn{1}{l}{}                                              & Precision      & Recall         & F1             & Precision      & Recall         & F1             & Precision      & Recall         & F1             & Precision      & Recall         & F1             \\ \hline
\cite{GhostLabeledRadarScenes} & -                     & 1.82 MB                              & 16.34                                                             & \textbf{92.58} & 90.59          & 91.57          & 94.92          & 93.91          & 94.41          & 99.78          & \textbf{99.92} & 99.85          & 95.76          & 94.80          & 95.28          \\ \hline
\cite{RadarMLPhDThesis}        & -                     & 1.82 MB                              & 18.23                                                             & 88.48          & 88.59          & 88.54          & 78.75          & 93.74          & 85.60          & 99.79          & 98.83          & 99.31          & 89.01          & 93.72          & 91.15          \\ \hline
GRU-TGTNet*                                 & -                     & 27.15 KB                             & 30.76                                                             & 87.90          & 88.49          & 88.19          & 94.25          & 93.36          & 93.80          & 99.82          & 99.84          & 99.83          & 93.99          & 93.90          & 93.94          \\ \hline
TGTNet-T (dim 16)                           & -                     & \textbf{11.52 KB}                    & 10.26                                                             & 88.39          & 89.83          & 89.10          & 94.72          & 93.78          & 94.25          & 99.85          & 99.83          & 99.84          & 94.32          & 94.48          & 94.40          \\ \hline
TGTNet-T (dim 32)                           & -                     & 40.91 KB                             & \textbf{9.94}                                                     & 90.21          & 91.26          & 90.73          & 95.66          & 94.68          & 95.17          & 99.86          & 99.87          & 99.87          & 95.24          & 95.27          & 95.26          \\ \hline
S-TGTNet-T (dim 32)                         & \xmark                     & 180.56 KB                            & 16.79                                                             & 91.64          & 92.13          & 91.89          & 95.95          & \textbf{95.49} & \textbf{95.72} & 99.88 & 99.88          & \textbf{99.88} & \textbf{95.82} & \textbf{95.83} & \textbf{95.83} \\ \hline
S-TGTNet-T (dim 32)                         & \cmark                     & 103.15 KB                            & 16.75                                                             & 91.45          & \textbf{92.35} & \textbf{91.90} & \textbf{96.05} & 94.91          & 95.48          & 99.86          & 99.88          & 99.87          & 95.79          & 95.71          & 95.75          \\ \hline
S-TGTNet-T\textsuperscript{\#} (dim 32)              & \cmark                                                                                         & 103.15 KB                            & 16.75                                                             & 90.45           & 91.90         & 91.17        & 95.74         & 95.37       & 95.55     & \textbf{99.89}            & 99.85        & 99.87       & 95.36          & 95.71      & 95.53     \\ \hline
SL-TGTNet-T (dim 16)                        & \cmark                     & 16.58 KB                             & 13.61                                                             & 89.20          & 91.05          & 90.11          & 95.74          & 91.30          & 93.46          & 99.73          & 99.88          & 99.80          & 94.89          & 94.07          & 94.46          \\ \hline
\end{tabular}
\end{threeparttable}
 }
 \begin{tablenotes}
     \item* reduced feature set
    \quad \revcolortext{\textsuperscript{\#} DBSCAN clustering}
     \quad \textsuperscript{\textdagger} Refers to Superpoint Block. All base models are LoRa.
 \end{tablenotes}
\label{table:results_temporal}
\end{table*}

\begin{table*}[!ht]
\caption{Comparison of our models on the Radar Ghost Dataset. Best values are shown in \textbf{bold}.}
\resizebox{\linewidth}{!}{
\begin{tabular}{cccccccccccccccc}
\hline
\multirow{2}{*}{\textbf{Setup}} & \multirow{2}{*}{\textbf{LoRa}} & \multirow{2}{*}{\textbf{Model size}} & \multirow{2}{*}{\textbf{\begin{tabular}[c]{@{}c@{}}Inf. time\\ (ms)\end{tabular}}} & \multicolumn{3}{c}{\textbf{Real}}                & \multicolumn{3}{c}{\textbf{Ghost}}               & \multicolumn{3}{c}{\textbf{Background}}          & \multicolumn{3}{c}{\textbf{Mean}}                \\
                                &                                &                                      &                                                                                    & Precision      & Recall         & F1             & Precision      & Recall         & F1             & Precision      & Recall         & F1             & Precision      & Recall         & F1             \\ \hline
PointNet\texttt{++}                      & -                              & 3.74 MB                              & 11.26                                                                              & 94.67          & 84.20          & 89.13          & 43.72          & 66.79          & 52.85          & \textbf{99.81} & \textbf{99.93} & \textbf{99.87} & 79.40          & 83.64          & 80.62          \\ \hline
TGTNet (dim 16)                 & \xmark          & 21.19 KB                             & 6.93                                                                      & 85.69          & 85.53          & 85.61          & 56.01          & 73.46          & 63.56          & 99.71          & 99.60          & 99.66          & 80.47          & 86.20          & 82.94          \\ \hline
TGTNet (dim 32)                  & \xmark          & 80.25 KB                             & \textbf{6.83}                                                                               & 89.82          & 81.09          & 85.23          & 54.00          & 70.39          & 61.11          & 99.61          & 99.77          & 99.69          & 81.14          & 83.75          & 82.01          \\ \hline
TGTNet (dim 16)                 & \cmark          & \textbf{11.21 KB}                    & 7.39                                                                               & \textbf{96.07} & 85.00          & 90.20          & 54.77          & \textbf{79.69} & 64.92          & 99.70          & 99.86          & 99.78          & 83.51          & 88.18          & 84.97          \\ \hline
S-TGTNet (dim 16)               & \cmark          & 28.03 KB                             & 12.01                                                                              & 94.87          & 80.26          & 86.96          & 52.66          & 68.14          & 59.41          & 99.55          & 99.88          & 99.71          & 82.36          & 82.76          & 82.03          \\ \hline
TGTNet-T (dim 16)                & \cmark          & 11.52 KB                             & 10.26                                                                              & 92.41          & \textbf{88.72} & 90.53          & 58.64          & 76.08          & 66.23          & 99.79          & 99.80          & 99.80          & 83.62          & \textbf{88.20} & 85.52          \\ \hline
S-TGTNet-T (dim 32)             & \cmark          & 103.15 KB                            & 16.75                                                                              & 95.44          & 86.16          & \textbf{90.57} & \textbf{61.28} & 72.60          & \textbf{66.46} & 99.69          & 99.89          & 99.79          & \textbf{85.47} & 86.22          & \textbf{85.61} \\ \hline
\end{tabular}
}
\vspace{-0.5cm}
\label{table:results_ghost_dataset}
\end{table*}

\subsubsection{Quantitative results}
The base TGTNet architecture is first evaluated against the GRT for different combinations of hyperparameters on the reduced feature set to analyze how the proposed size reduction techniques affect performance, with results reported in Table \ref{table:results_base}. It can be observed that using LoRa layers instead of regular layers has a negligible effect on performance while simultaneously halving the memory footprint. However, despite LoRa layers requiring fewer operations for the matrix multiplications, there is, on average, a less than 1ms (5\%) increase in inference time due to the matrix multiplications executing sequentially. A greater performance difference is observed when changing the model size directly. It can be seen that the performance degrades by only 1--2\% across metrics when using the smaller model (dim 16 vs. dim 32), while the memory footprint is four times smaller. Such a performance drop can be justifiable when resources are scarce. Compared to the GRT, the best-performing TGTNet models only have a 2\% loss in moving object F1 performance and a 1\% loss in clutter F1 performance, while being 200 times smaller and having almost a 4 times faster inference time.

Table \ref{table:results_superpoints} shows tests for Superpoint block augmentations using LoRa models as the base. It can be noticed that the regular Superpoint Block generally performs better than the light Superpoint Block. SL-TGTNet achieves only a 1\% improvement for moving object F1 compared to TGTNet, while S-TGTNet achieves a 4\% moving object F1 increase and a 1\% clutter F1 increase. This occurs because the normal Superpoint Block provides an additional full GTB that adds expressivity to the model. By testing LoRa variants of the Superpoint Blocks independently from the base model, we observe that LoRa offers no performance degradation. Regarding the model size, an increase of 1\% F1 for moving objects and an almost 2\% increase in F1 for clutter objects can be noticed, increases which are similar to those observed in the base models. 

Additional tests are done by augmenting the base model with different methods that embed temporal information into the model, as can be seen in Table \ref{table:results_temporal}. The first method, GRU-TGTNet, involves adding a graph GRU module as a temporal feature extraction block before the model. The second method, TGTNet-T, retains the same parameter set as the base model and utilizes the downsampling layer as a shared layer across different frames before aggregating them together. It can be noticed that, despite its simplicity, TGTNet-T achieves 1\% higher clutter and moving object F1 scores compared to the more complex and slower GRU feature extraction method.

Further performance can be achieved by combining Superpoint Block feature extraction with parallel downsampling, as can be observed from the results of the TGTNet-T models. Every frame is processed separately by the same shared Superpoint Block and then passed onto the parallel downsampling network. This method offers an advantage of more than 1\% F1 for the clutter class and achieves the same or higher F1 for moving objects, compared to the best metrics obtained by the baselines, while having a 20 times smaller memory footprint, and maintaining similar inference time. Moreover, it can be observed that the regular Superpoint Block brings a 1\% increase in F1 for moving objects in the temporal model, while the clutter detection performance is similar.

\revcolortext{Tables ~\ref{table:results_superpoints} and ~\ref{table:results_temporal} also feature comparisons of DBSCAN with energy model clustering, demonstrating that the latter yields superior mean F1 scores across temporal and non-temporal model variants without increasing computational cost.} 

Table \ref{table:results_ghost_dataset} shows results for our model against a modified version of PointNet\texttt{++}. This version keeps the exact same architecture as the ``semantic scene labelling'' architecture described in the original PointNet\texttt{++} paper\cite{PointNet++}, with downsampling set to 0.5. It can be noticed that our smallest TGTNet model outperforms PointNet\texttt{++} by almost 9\% in Real object F1 and 20\% in ghost F1. Surprisingly, modifications that increase the model's complexity appear to reduce its performance. We argue that this happens because of the dataset's small size, which encourages overfitting.

\subsubsection{Qualitative results}

\begin{figure*}[!ht]
\centering

\begin{subfigure}[t]{0.49\linewidth}
    \includegraphics[width=\linewidth]{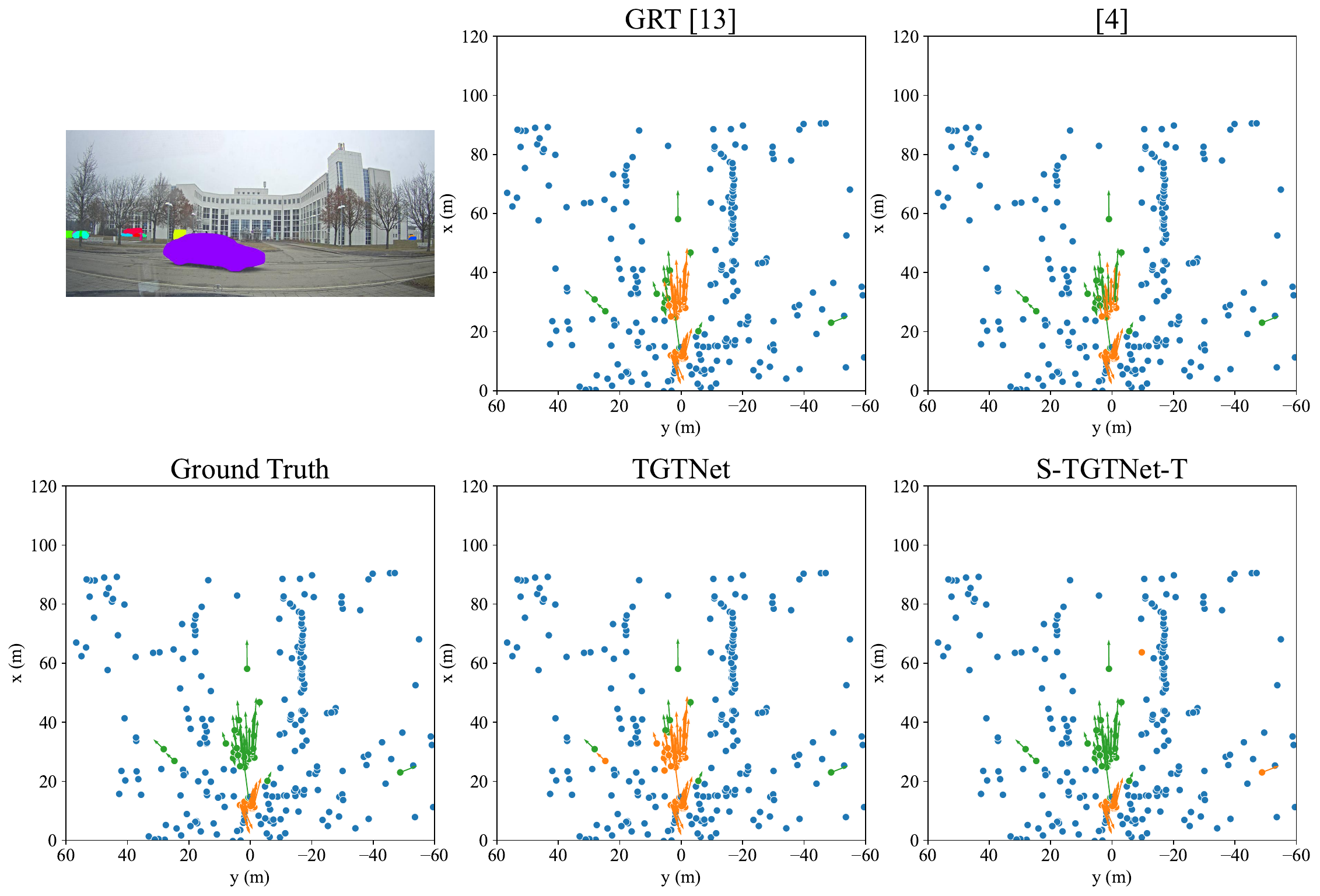}
    \caption{}
    \label{fig:side_by_side-a}
\end{subfigure}
\begin{subfigure}[t]{0.49\linewidth}
    \includegraphics[width=\linewidth]{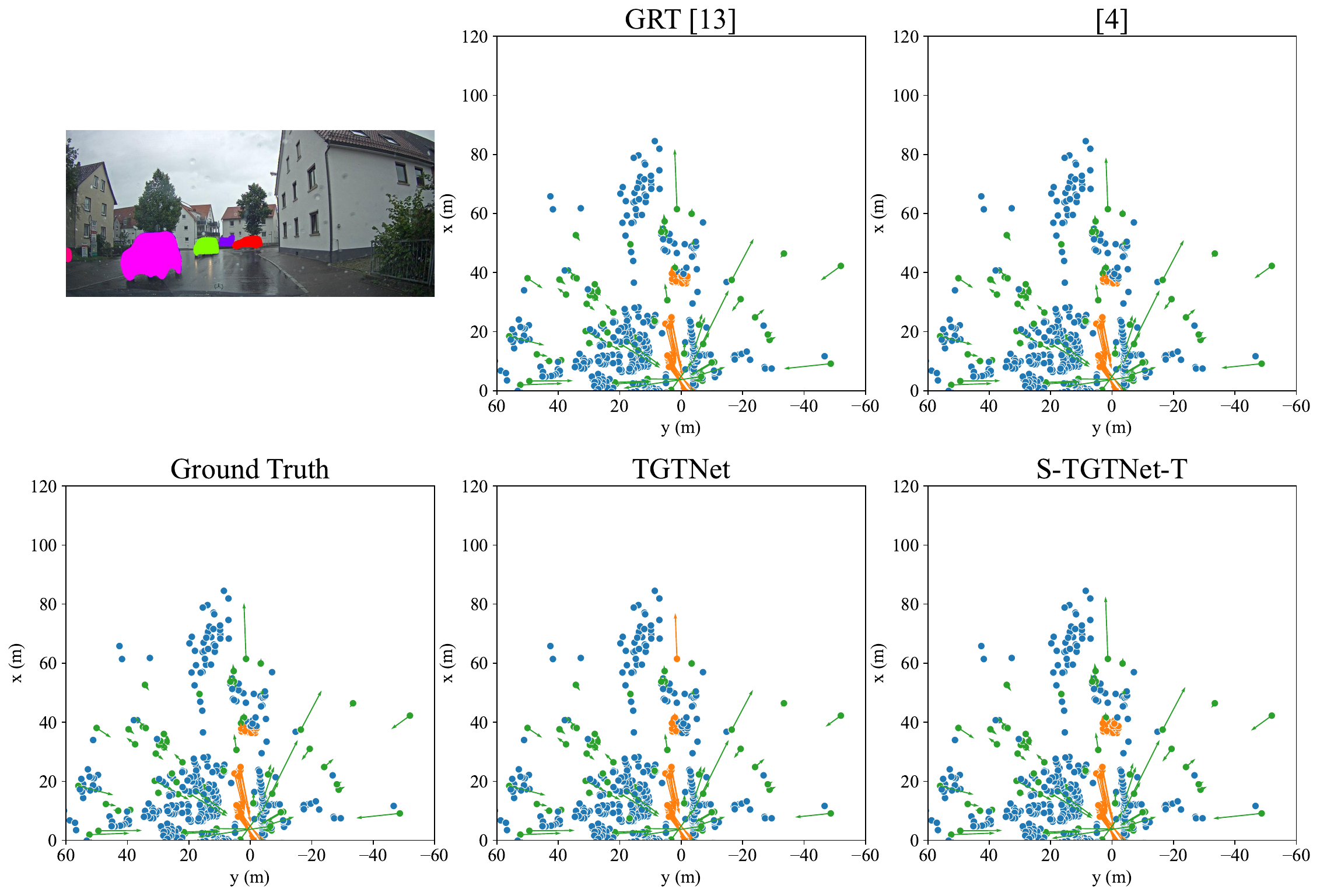}
    \caption{}
    \label{fig:side_by_side-b}
\end{subfigure}

\caption{Comparison of our models against baselines for two scenes (\subref{fig:side_by_side-a}) and (\subref{fig:side_by_side-b}). Stationary detections are shown in blue \mycirc[stationary-blue], moving object detections in orange \mycirc[moving-object-orange], and clutter in green \mycirc[clutter-green]. TGTNet is chosen as the LoRa variant with dimension 16, and S-TGTNet-T is chosen as the LoRa variant with dimension 32 and extra input features.\vspace{-0.5cm}}
\label{fig:side_by_side}

\end{figure*}

For qualitative results, we compare the two best-performing baselines, the GRT \cite{GaussianRadarTransformer} and \cite{GhostLabeledRadarScenes} against our smallest model, the LoRa dim 16 TGTNet, and one of the best-performing models, the LoRa dim 32 S-TGTNet-T.

As shown in Fig. \ref{fig:side_by_side} (\subref{fig:side_by_side-a}), S-TGTNet-T accurately classifies all structured clutter detections, unlike the larger models. The smaller 11KB TGTNet variant is also close to the larger models in its classification.

In Fig. \ref{fig:side_by_side} (\subref{fig:side_by_side-b}), the detections surrounding the vehicle marked in the image in red are ambiguous. Our smaller model, TGTNet, incorrectly estimates the spatial structure of the detections while also omitting clutter detections. S-TGTNet-T, however, correctly predicts all types of detections and their shapes, but is more conservative with its stationary and clutter prediction, thus estimating a larger moving object footprint than the ground truth and the larger models.

\section{Conclusions and future work}
\label{sect:conc}

We presented lightweight graph transformer models for clutter and target classification in radar point clouds. Different lightweight variants of the proposed TGTNet model, along with different temporal augmentations, were developed and evaluated. Among these models, applying parallel downsampling and merging the downsampled point clouds was found to be particularly effective, despite its simplicity.
Adding superpoint feature extraction blocks further kept the model size small while achieving precision, recall and F1 classification scores compared to the much larger baseline models.




Finally, based on the performance that temporal augmentations bring, future work could involve utilizing pre-trained point cloud flow estimator networks to estimate the past positions of the points in the current point cloud, yielding more informative neighborhoods. Since the proposed models use 32-bit floating-point parameters,  methods such as quantization-aware training could prove useful in further reducing memory footprint and computational cost. Future work will also focus on further optimizing the proposed models for deployment on automotive-grade processors.

\bibliographystyle{IEEETran}
\bibliography{egbib}

\begin{thebibliography}{10}
\providecommand{\url}[1]{#1}
\csname url@samestyle\endcsname
\providecommand{\newblock}{\relax}
\providecommand{\bibinfo}[2]{#2}
\providecommand{\BIBentrySTDinterwordspacing}{\spaceskip=0pt\relax}
\providecommand{\BIBentryALTinterwordstretchfactor}{4}
\providecommand{\BIBentryALTinterwordspacing}{\spaceskip=\fontdimen2\font plus
\BIBentryALTinterwordstretchfactor\fontdimen3\font minus \fontdimen4\font\relax}
\providecommand{\BIBforeignlanguage}[2]{{%
\expandafter\ifx\csname l@#1\endcsname\relax
\typeout{** WARNING: IEEEtran.bst: No hyphenation pattern has been}%
\typeout{** loaded for the language `#1'. Using the pattern for}%
\typeout{** the default language instead.}%
\else
\language=\csname l@#1\endcsname
\fi
#2}}
\providecommand{\BIBdecl}{\relax}
\BIBdecl

\bibitem{RuleBasedClutterDetection}
J.~Kopp, D.~Kellner, A.~Piroli, and K.~Dietmayer, ``Fast rule-based clutter detection in automotive radar data,'' in \emph{2021 IEEE Intl. Intelligent Transportation Systems Conf. (ITSC)}.\hskip 1em plus 0.5em minus 0.4em\relax IEEE, 2021, pp. 3010--3017.

\bibitem{GhostDataset}
F.~Kraus, N.~Scheiner, W.~Ritter, and K.~Dietmayer, ``The radar ghost dataset--an evaluation of ghost objects in automotive radar data,'' in \emph{IEEE Intl. Conf. Intelligent Robots and Systems}.\hskip 1em plus 0.5em minus 0.4em\relax IEEE, 2021, pp. 8570--8577.

\bibitem{RadarScenes}
O.~Schumann, M.~Hahn \emph{et~al.}, ``{RadarScenes: A real-world radar point cloud data set for automotive applications},'' in \emph{2021 IEEE 24th Intl. Conf. Information Fusion (FUSION)}.\hskip 1em plus 0.5em minus 0.4em\relax IEEE, 2021, pp. 1--8.

\bibitem{GhostLabeledRadarScenes}
J.~Kopp, D.~Kellner, A.~Piroli, and K.~Dietmayer, ``{Tackling clutter in radar data-label generation and detection using PointNet++},'' in \emph{IEEE Intl. Conf. on Robotics and Automation}.\hskip 1em plus 0.5em minus 0.4em\relax IEEE, 2023, pp. 1493--1499.

\bibitem{RSB}
F.~Roos, M.~Sadeghi, J.~Bechter, N.~Appenrodt, J.~Dickmann, and C.~Waldschmidt, ``{Ghost target identification by analysis of the Doppler distribution in automotive scenarios},'' in \emph{2017 18th International Radar Symposium (IRS)}, 2017, pp. 1--9.

\bibitem{ZLL}
L.~Zheng, J.~Long, M.~Lops, F.~Liu, X.~Hu, and C.~Zhao, ``Detection of ghost targets for automotive radar in the presence of multipath,'' \emph{IEEE Transactions on Signal Processing}, vol.~72, pp. 2204--2220, 2024.

\bibitem{DB}
S.~Danino and I.~Bilik, ``Automatic multipath annotation for conventional automotive radar datasets,'' \emph{IEEE Sensors Journal}, vol.~24, no.~8, pp. 13\,500--13\,517, 2024.

\bibitem{PointNet++}
C.~R. Qi, L.~Yi, H.~Su, and L.~J. Guibas, ``{PointNet++: Deep hierarchical feature learning on point sets in a metric space},'' \emph{Advances in Neural Information Processing Systems}, vol.~30, 2017.

\bibitem{SGPN}
W.~Wang, R.~Yu, Q.~Huang, and U.~Neumann, ``{SGPN: Similarity group proposal network for 3d point cloud instance segmentation},'' in \emph{Proc. of IEEE Conf. Comp. Vision \& Pattern Recognition}, 2018, pp. 2569--2578.

\bibitem{DBSCAN}
M.~Ester, H.-P. Kriegel, J.~Sander, X.~Xu \emph{et~al.}, ``A density-based algorithm for discovering clusters in large spatial databases with noise,'' in \emph{kdd}, vol.~96, 1996, pp. 226--231.

\bibitem{RadarMLPhDThesis}
O.~Schumann, ``Machine learning applied to radar data: classification and semantic instance segmentation of moving road users,'' Ph.D. dissertation, Dissertation, Dortmund, Technische Universit{\"a}t, 2021.

\bibitem{PointTransformer}
H.~Zhao, L.~Jiang, J.~Jia, P.~H. Torr, and V.~Koltun, ``Point transformer,'' in \emph{Proc. of the IEEE Intl. Conf. Comp. Vision}, 2021, pp. 16\,259--16\,268.

\bibitem{GaussianRadarTransformer}
M.~Zeller, J.~Behley, M.~Heidingsfeld, and C.~Stachniss, ``Gaussian radar transformer for semantic segmentation in noisy radar data,'' \emph{IEEE Robotics and Automation Letters}, vol.~8, no.~1, pp. 344--351, 2022.

\bibitem{gheorghiu2026tgtnet}
A.~Gheorghiu, T.~Alkanat, and A.~Pandharipande, ``Graph transformers for automotive radar clutter and target classification at the edge,'' in \emph{IEEE International Conference on Acoustics, Speech \& Signal Processing (ICASSP)}, 2026, to appear.

\bibitem{GELU}
D.~Hendrycks and K.~Gimpel, ``Gaussian error linear units (gelus),'' \emph{arXiv preprint arXiv:1606.08415}, 2016.

\bibitem{BatchNorm}
S.~Ioffe and C.~Szegedy, ``Batch normalization: accelerating deep network training by reducing internal covariate shift,'' in \emph{Proc. 32nd Intl. Conf. Machine Learning - Volume 37}.\hskip 1em plus 0.5em minus 0.4em\relax JMLR.org, 2015, p. 448–456.

\bibitem{SuperpointGraphs}
L.~Landrieu and M.~Simonovsky, ``Large-scale point cloud semantic segmentation with superpoint graphs,'' in \emph{2018 IEEE/CVF Conference on Computer Vision and Pattern Recognition}, 2018, pp. 4558--4567.

\bibitem{cutpursuit}
L.~Landrieu and G.~Obozinski, ``Cut pursuit: Fast algorithms to learn piecewise constant functions on general weighted graphs,'' \emph{SIAM Journal on Imaging Sciences}, vol.~10, no.~4, pp. 1724--1766, 2017.

\bibitem{UNet}
O.~Ronneberger, P.~Fischer, and T.~Brox, ``U-net: Convolutional networks for biomedical image segmentation,'' in \emph{Medical Image Computing and Computer-Assisted Intervention -- MICCAI 2015}, N.~Navab, J.~Hornegger, W.~M. Wells, and A.~F. Frangi, Eds.\hskip 1em plus 0.5em minus 0.4em\relax Cham: Springer International Publishing, 2015, pp. 234--241.

\bibitem{SimplifiedPointAttention}
B.~Yang, S.~Wang, A.~Markham, and N.~Trigoni, ``Robust attentional aggregation of deep feature sets for multi-view 3d reconstruction,'' \emph{Intl. Journal of Computer Vision}, vol. 128, no.~1, pp. 53--73, 2020.

\bibitem{LoRa}
E.~J. Hu, yelong shen, P.~Wallis, Z.~Allen-Zhu, Y.~Li, S.~Wang, L.~Wang, and W.~Chen, ``Lo{RA}: Low-rank adaptation of large language models,'' in \emph{International Conference on Learning Representations}, 2022.

\bibitem{DeepSeek}
A.~Liu, B.~Feng, B.~Xue, B.~Wang, B.~Wu, C.~Lu, C.~Zhao, C.~Deng, C.~Zhang, C.~Ruan \emph{et~al.}, ``Deepseek-v3 technical report,'' \emph{arXiv preprint arXiv:2412.19437}, 2024.

\bibitem{pointrnn}
H.~Fan and Y.~Yang, ``{PointRNN: Point recurrent neural network for moving point cloud processing},'' \emph{arXiv preprint arXiv:1910.08287}, 2019.

\bibitem{LovaszLoss}
M.~Berman, A.~R. Triki, and M.~B. Blaschko, ``The lov{\'a}sz-softmax loss: A tractable surrogate for the optimization of the intersection-over-union measure in neural networks,'' in \emph{Proceedings of the IEEE conference on computer vision and pattern recognition}, 2018, pp. 4413--4421.

\bibitem{BalancedLoss}
Y.~Cui, M.~Jia, T.-Y. Lin, Y.~Song, and S.~Belongie, ``Class-balanced loss based on effective number of samples,'' in \emph{Proc. of the IEEE/CVF Conf. on Computer Vision and Pattern Recognition}, 2019, pp. 9268--9277.

\bibitem{adam}
D.~P. Kingma and J.~Ba, ``Adam: {A} method for stochastic optimization,'' in \emph{3rd International Conference on Learning Representations, {ICLR} 2015, San Diego, CA, USA, May 7-9, 2015, Conference Track Proceedings}, Y.~Bengio and Y.~LeCun, Eds., 2015.

\bibitem{CosineAnnealing}
I.~Loshchilov and F.~Hutter, ``{SGDR: Stochastic Gradient Descent with Warm Restarts},'' in \emph{ICLR (Poster)}, 2017.

\bibitem{ADASDeepLearning}
F.~Dellinger, T.~Boulay, D.~M. Barrenechea, S.~El-Hachimi, I.~Leang, and F.~B{\"u}rger, ``{Multi-task network pruning and embedded optimization for real-time deployment in ADAS},'' \emph{preprint arXiv:2101.07831}, 2021.

\end{thebibliography}

\end{document}